\documentclass[final]{cvpr}

\usepackage{times}
\usepackage{epsfig}
\usepackage{graphicx}
\usepackage{comment}
\usepackage{amsmath}
\usepackage{amssymb}

\usepackage[pagebackref=true,breaklinks=true,colorlinks,bookmarks=false]{hyperref}

\usepackage{multirow}

\usepackage[dvipsnames]{xcolor}

\usepackage{sidecap}

\usepackage[labelsep=quad,indention=10pt]{subfig}
 \newcommand{\gs}[1]{}
 \newcommand{\kl}[1]{}
 \newcommand{\bs}[1]{}

\newcommand{\kledit}[1]{{#1}}
\newcommand{\bsedit}[1]{{#1}}

\def\v#1{\mathbf{#1}}
\DeclareMathOperator*{\argmax}{arg\,max}

\def\cvprPaperID{10239} 
\def\confYear{CVPR 2021}

\begin{document}

\title{Fingerspelling Detection in American Sign Language}

\author{Bowen Shi$^1$, Diane Brentari$^2$, Greg Shakhnarovich$^1$, Karen Livescu$^1$\\
$^1$Toyota Technological Institute at Chicago, USA $^2$University of Chicago, USA\\
{\tt\small \{bshi,greg,klivescu\}@ttic.edu \ \ \ \ \ dbrentari@uchicago.edu}
}

\twocolumn[{%
\renewcommand\twocolumn[1][]{#1}%
\maketitle
\begin{center}
\vspace{-0.75cm}
\includegraphics[width=\textwidth]{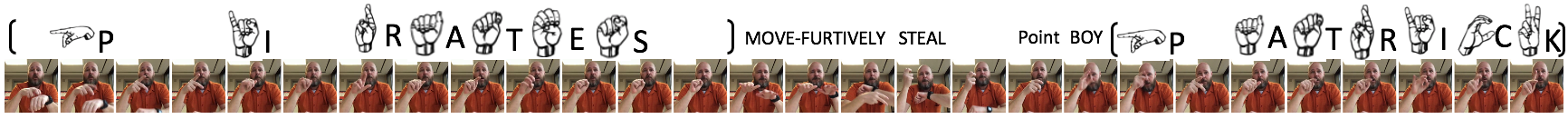}
  \captionof{figure}{\label{fig:teaser}Fingerspelling detection and recognition in video of American Sign Language.
    The goal of \emph{detection} is to find
    intervals corresponding to fingerspelling (here indicated by
  open/close parentheses), and the goal of \emph{recognition} is to
  transcribe each of those \kledit{intervals} into letter sequences. Our focus in
  this paper is on detection that enables accurate recognition.  In this example (with downsampled frames), 
the fingerspelled words are PIRATES and PATRICK, shown along with their canonical handshapes \kledit{aligned roughly with the most-canonical corresponding frames}.  Non-fingerspelled signs are labeled with their glosses.  The English translation is ``Moving furtively, pirates steal the boy Patrick.''}
\end{center}
}]


\begin{abstract}
Fingerspelling, in which words are signed letter by letter, is an
important component of American Sign Language. \kledit{Most} previous work on
automatic fingerspelling recognition has assumed that the boundaries of
fingerspelling regions in signing videos are known beforehand. In this
paper, we consider the task of fingerspelling detection
in raw, untrimmed sign language videos.  This is an important step towards building real-world fingerspelling recognition systems. 
We
propose a benchmark \kledit{and a suite of evaluation metrics, some of which reflect the effect
of detection on the downstream fingerspelling
recognition task.} In addition, we propose a new model that learns \kledit{to detect fingerspelling} via multi-task training, incorporating pose estimation and
fingerspelling recognition (transcription) along with \kledit{detection, and compare this model to several alternatives}. The model outperforms all alternative approaches across all
metrics, establishing a state of the art on the benchmark.
\end{abstract}

\section{Introduction}
Sign languages, such as American Sign Language (ASL), are
natural languages expressed via
movements of the hands, face, and upper body.
Automatic processing of sign languages would assist communication between deaf and hearing
individuals,
but involves
a number of challenges. There is no standard written form for sign
languages.
\kledit{Automatic transcription} of sign language into
a written language such as English is in general a translation task. In addition,
sign language gestures are often coarticulated and do
not appear in their canonical forms~\cite{jerde2003coarticulation,Keane2012ac}.

In this paper, we focus on fingerspelling \kledit{(Figure~\ref{fig:teaser})}, a component of sign
language in which words are signed letter by letter, with a
distinct handshape or trajectory corresponding to each letter in the
alphabet of a written language (e.g., the English alphabet for ASL fingerspelling).
Fingerspelling is used for \kledit{multiple} purposes,
including for words that do not have their own signs (such as many
proper nouns, technical terms, and
abbreviations)~\cite{padden1998asl} but also sometimes for emphasis or
expediency. 
%
Fingerspelling accounts for 12\% to 35\% of ASL, where it is used more than in other sign languages~\cite{padden2013alphabet}. As important content words are
commonly fingerspelled, automatic fingerspelling recognition can
enable practical tasks such as search and retrieval in ASL media.

Compared to \emph{translation}
between
\kledit{a sign language} and a written language, fingerspelling recognition involves
\emph{transcription} into a  restricted set of symbols, with a monotonic alignment with the written form. Linguistically, fingerspelling is distinct from other elements of ASL,
such as
lexical
signs and classifiers~\cite{johnston1999defining,brentari2001language},
so \kledit{fingerspelling is likely to benefit from a separate model.}
The role that fingerspelling transcription is likely to play in ASL to English translation is
  similar to that of transliteration in written language
  translation~\cite{durrani2014integrating}. For all of these reasons, we believe that even as more general ASL
  \kledit{processing} methods are developed,
  \kledit{it will continue to be beneficial to have} dedicated fingerspelling detection and recognition
modules. 

Fingerspelling recognition
has been widely
studied~\cite{pugeault,goh,ricco,liw,tkim3,shi2019iccv}.
However, in \kledit{most} prior work, it is assumed that the
input sequence contains fingerspelling \kledit{only, sometimes} extracted from longer sequences of signing via human annotation.
Replacing human annotation with fully automatic
\emph{detection} of fingerspelling -- identifying time spans in the
video containing fingerspelling -- is a hurdle that
must be cleared to enable truly practical fingerspelling
recognition ``in the wild''.

Fingerspelling detection has not been \kledit{widely} studied before. 
\kledit{In principle}
  it can be treated as a special
case of action
detection~\cite{xu2017rc3d,chao2018rethinking,zhao2017temporal,Escorcia2016DAPsDA}. However,
in contrast to typical action detection scenarios, the actions in the
fingerspelling ``class'' are highly \bsedit{heterogeneous} and many fingerspelling handshapes are also used in non-fingerspelled signs.  In addition, considering the
goal of using the detector as part of a complete sign language processing system, a fingerspelling detector should be evaluated based on its effect on a downstream recognition
model, a step not normally
included in evaluation of action recognition. This makes common
detection metrics, like average precision (AP) for action detection, less informative for fingerspelling
detection.

Our design of a detection model is motivated by two
observations. The first is that articulated pose, in particular
handshape, plays a role in the distinctiveness of fingerspelling from other types of sign. At the same time, pose
estimation, while increasingly successful in some domains, may be
insufficiently accurate for directly informing fingerspelling
recognition, as shown in~\cite{shi2019iccv} and in our experiments. Instead we
incorporate pose estimation as part of
 training 
our model, but do not rely on explicit pose estimates at test time.
The second observation concerns the goal of optimizing 
fingerspelling detection as a means to an end of improving downstream
recognition. We address this by including a fingerspelling recognizer
in model training. Our results show that this multi-task learning
approach produces a superior detector compared to baselines that omit
the pose and/or recognition losses.

Ours is to our knowledge the first work to demonstrate the effect of fingerspelling detection on
fully automatic fingerspelling recognition in the wild.
Beyond this novelty, our contributions are as follows. First, we propose an
evaluation framework for fingerspelling detection that incorporates
the downstream recognition task into the metrics,
and introduce a benchmark based on extending a publicly available data set.
Second, we investigate a number of approaches for fingerspelling
detection, \bsedit{adapted} from fingerspelling recognition and action
detection, and develop a novel multi-task learning approach.
Our model \bsedit{outperforms} baseline detection approaches across all
evaluation metrics, establishing a state of the art for the proposed
benchmark.

\section{Related Work}

Early work on sign language
recognition
mostly
focused on recognition of isolated signs. Recent work has increasingly focused on continuous sign language
recognition, which transforms image sequences to glosses (for general sign language) or letter sequences (for fingerspelling), \kledit{and translation}. 
Approaches commonly separate between an image feature extraction component
and a sequence modeling component. The former has moved from employing
human-engineered features such as HoG~\cite{tkim3,koller15cslr} to
convolutional
networks~\cite{koller16deephand,koller16hybridsign,shi2018slt}. 
Approaches for the sequence modeling component have included Hidden Markov Models
(HMM)~\cite{koller16hybridsign,koller2017resign,koller2019multistream}, semi-Markov conditional random fields~\cite{tkim3}, 
connectionist temporal classification
(CTC)~\cite{shi2018slt,shi2019iccv,Cui2017RecurrentCN}, and
\kledit{encoder-decoder models~\cite{Huang2018VideobasedSL,camgoz2020sign}}, which are
largely borrowed from speech recognition.

Much of the prior work has been limited to data collected in a controlled
environment.
There has been a
growing interest in sign language recognition ``in the wild'' (naturally occurring \bsedit{sign language} media), which includes challenging visual conditions such as lighting variation, visual clutter, and motion blur, and often also more natural signing styles~\cite{Forster2012RWTHPHOENIXWeatherAL,Forster2014ExtensionsOT,albanie2020eccv,Joze2019MSASLAL,li2020word,li2020transferring,shi2018slt,shi2019iccv}.
Two recently released datasets of fingerspelling in the wild~\cite{shi2018slt,shi2019iccv} include 
data from 168 signers and tens of thousands of fingerspelling segments; these are the testbeds used in our
experiments.

Prior work on fingerspelling detection~\cite{yanovich2016detection,tsechpenakis2006robust,tsechpenakis2006learning,yang2010simultaneous} employs visual features from optical
flow or pre-defined hand keypoints.  \kledit{For sign language data collected in the wild, the quality of pose estimates is usually low, making them a poor choice as input to a detector (as we show in our experiments).}
Tsechpenakis \emph{et al.}~\cite{tsechpenakis2006robust,tsechpenakis2006learning}
proposed a statistical procedure to detect changes in video for fingerspelling detection; however, these were tested anecdotally, in controlled environments and simplified scenarios.
Related work on detection of sign language segments (in video containing both signing and non-signing) uses recurrent neural networks (RNN) to classify individual frames into signing/non-signing categories~\cite{Moryossef2020sld}.  \kledit{We compare our approach to baselines that use ideas from this prior work.} 

Prior work~\cite{Moryossef2020sld,yanovich2016detection} has largely
treated the detection task as frame classification, evaluated via classification accuracy.  Though sequence prediction is considered in~\cite{tsechpenakis2006robust,tsechpenakis2006learning}, the model evaluation is qualitative.  In practice, a detection model is often intended to serve as one part of a larger system, producing candidate segments to a downstream recognizer.  Frame-based metrics ignore the quality of the segments and their effect on a downstream recognizer.

\kledit{Our modeling approach is based on the intuition that training
  with related tasks including recognition and pose estimation should
  help in detecting fingerspelling segments.
Multi-task approaches have been studied for other related tasks. For example, Li {\it et al.}~\cite{Li2017TowardsET} jointly train an object detector and a word recognizer to perform text spotting. In contrast to this approach of treating detection and recognition as two parallel tasks, our approach further allows the detector to account for the performance of the recognizer. In sign language recognition, Zhou {\it et al.}~\cite{Zhou2020SpatialTemporalMN} estimate human pose keypoints while training the recognizer.  However, the keypoints used in~\cite{Zhou2020SpatialTemporalMN} are manually annotated. Here we study whether we can distill knowledge from an external imperfect pose estimation model (OpenPose).}

\begin{figure*}[tb]
  \centering
  \includegraphics[width=\textwidth]{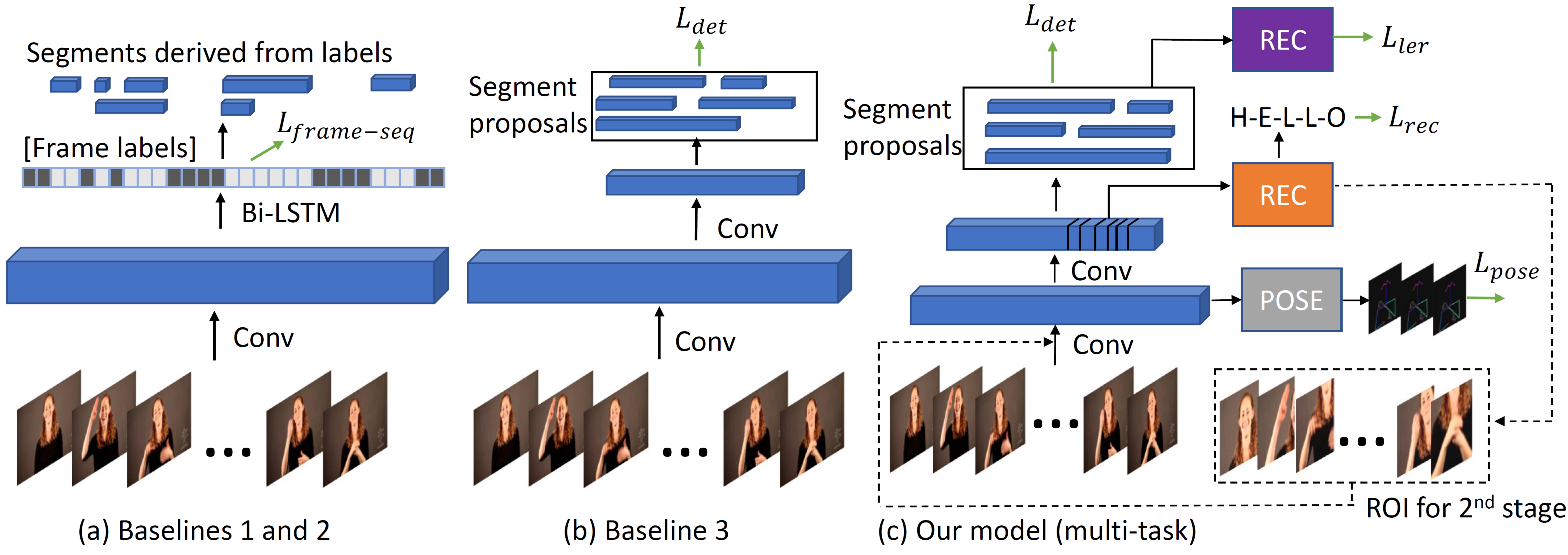}
  \caption{\label{fig:model_arch} Detection models. (a) Baselines 1
    and 2. \kledit{The frame labels are binary labels and augmented
    letter labels, respectively,} for baselines 1 and 2. $L_{frame-seq}$
    is \kledit{a per-frame cross-entropy loss} for baseline 1 and modified CTC
    loss for baseline 2. (b): Baseline 3, modified R-C3D; $L_{det}$ is
    the standard detection loss. (c) Our multi-task model trained with a
    combination of detection loss, \kledit{recognition loss using a
    recognizer that takes in the ground-truth segments; letter error loss using a
    recognizer that takes in the detections;} and pose estimation loss computed relative to pose
    \kledit{pseudo-labels.}}
\end{figure*}

\section{Task and metrics}
\label{sec:metrics}

\newcommand{\letseq}{\mathbf{l}}

We are given a sign language video clip with $N$
frames $I_1,..., I_N$ containing $n$
fingerspelling segments \kledit{$\{x^{\ast}_i\}_{1\leq i\leq n}$} ($\ast$ denotes ground truth), where
$x^{\ast}_i=(s^{\ast}_i, t^{\ast}_i)$ 
  and $s^{\ast}_i, t^{\ast}_i$ are the start and end frame indices of the $i^\textrm{th}$ segment. \kledit{The corresponding ground-truth letter sequences are $\{\letseq_i^\ast\}_{1\leq i\leq n}$.}
  \textbf{Detection task} The task of fingerspelling
  detection is to find the fingerspelling segments within the clip. A detection model outputs $m$ predicted segment-score pairs $\{(\widehat{x}_i, {f}_i)\}_{1\leq i\leq m}$ where $\widehat{x}_i$ and $f_i$ are the $i^\text{th}$ predicted segment and its confidence score, respectively.

  \textbf{AP@IoU} A metric commonly used in object
detection~\cite{lin2014microsoft} and action
detection~\cite{idrees2016thumos,caba2015activitynet} is AP@IoU. Predicted
segments, sorted by score $f_i$, are sequentially matched to the
ground-truth segment $x^{\ast}_j$ with the highest IoU (intersection
over union, a measure of overlap) above a
threshold $\delta_{IoU}$. Once $x^\ast_j$ is matched to a
$\widehat{x}_i$ it is removed from the candidate set.  Let
$k(i)$ be the index of the ground-truth segment $x^\ast_{k(i)}$ matched to $\widehat{x}_i$; then formally,
\vspace{-.05in}
\begin{equation}
  \label{eq:ap_iou}
  k(i)=\displaystyle\argmax_{j: IoU(\widehat{x}_i,
    x^{\ast}_j)>\delta_{IoU},\,j\ne k(t)\forall t<i}IoU(\widehat{x}_i, x^{\ast}_j).
\end{equation}
Precision and recall are defined as the proportions of matched
examples in the \kledit{predicted and ground-truth segment sets},
respectively. Varying the number of predictions $m$ gives a \kledit{precision-recall curve $p(\tilde{r})$}.  The average precision (AP) is defined as the mean precision \kledit{over} $N_r+1$ equally spaced recall levels $[0, 1/N_r,...,1]$\footnote{$N_r$ is set to 100 as in~\cite{lin2014microsoft}.}, \kledit{where} the precision \kledit{at a given recall is defined as}
the maximum precision at a recall exceeding that level:
$AP=\frac{1}{N_r}\displaystyle\sum_{i=1}^{N_r}\max_{\tilde{r}:\tilde{r}\geq
  i/N_r}p(\tilde{r})$. AP@IoU can be reported for a range of values of
$\delta_{IoU}$.

\newcommand{\acc}{\textrm{Acc}}

\textbf{Recognition task} The task of recognition is to transcribe $\widehat{x}=(\widehat{s},\widehat{t})$ into a letter
sequence $\widehat{\letseq}$. The recognition accuracy of a predicted sequence
$\widehat{\letseq}$ w.r.t.~the ground truth $\letseq^\ast$ is defined as
$\acc(\letseq^\ast,\widehat{\letseq}) = 1 -
D(\letseq^\ast,\widehat{\letseq})/|\letseq^\ast|$, where $D$ is the
edit (Levenstein) distance
and $|\letseq|$ is
the length of a sequence. Note that $\acc$ can be negative. 

Prior work has considered recognition mainly applied to ground-truth segments $x^\ast$; in contrast, here we are concerned
with \kledit{detection for the purpose of recognition}.
We match a predicted $\widehat{x}_j$ to a ground-truth
$x^{\ast}_i$\bs{}
and
then evaluate the accuracy of $\widehat{\letseq}_j$
w.r.t.~$\letseq_i^\ast$. Thus, in contrast to a typical action
detection scenario, here
IoU may not be
perfectly correlated with recognition accuracy. For example, a
detected segment that is too short can hurt recognition much more than
a segment that is too long.
We propose a new metric, AP@Acc, to measure the
performance of a fingerspelling detector in the context of a
given downstream recognizer. 

\textbf{AP@Acc} 
This metric uses the letter accuracy of a
recognizer to match between predictions and ground-truth. It also
requires \kledit{an IoU threshold} to prevent matches between non-overlapping
segments. As in AP@IoU, predicted segments are sorted by score and
sequentially matched:
\vspace{-.05in}
\begin{equation}
  \label{eq:ap_acc}
    k(i)=\displaystyle\argmax_{\substack{j: IoU(\widehat{x}_i,
        x^{\ast}_j)>\delta_{IoU},\;j\ne k(t)\,\forall t<i\\
        Acc(\letseq^\ast_j,Rec(I_{\widehat{s}_i:\widehat{t}_i}))>\delta_{acc}}}
    \hspace{-2em}Acc(\letseq_j^\ast,Rec(I_{\widehat{s}_i:\widehat{t}_i})),
\end{equation}
where $Rec(I_{s:t})$ is the output (predicted letter sequence) from a
recognizer given the frames $I_{s:t}$.  We can
report AP@Acc for multiple values of $\delta_{acc}$.

%

\textbf{Maximum Sequence Accuracy (MSA)} Both AP@IoU and AP@Acc measure the precision of a set of detector predictions.
Our last metric directly measures just the performance of a
given downstream recognizer when given the detector's
predictions. We form the ground-truth letter sequence for the entire
video $I_{1:N}$ by concatenating the
letters of all ground-truth segments, with a special ``no-letter''
symbol $\emptyset$ separating consecutive letter sequences:
\vspace{-.05in}
\begin{equation}
  \mathbf{L}^\ast=\emptyset,\letseq^\ast_1,\emptyset,\ldots,\emptyset,\letseq^\ast_n,\emptyset.
  \label{eq:lstar}
\end{equation}
\bsedit{Note that $\emptyset$ is inserted only where non-fingerspelling frames exist.} For instance, the video in Figure~\ref{fig:teaser} would yield
$\mathbf{L}^\ast=\text{PIRATES}\emptyset\text{PATRICK}$.
We similarly obtain a full-video letter sequence from the predicted
segments. We suppress detections with score $f_i$ below $\delta_f$ and apply local
non-maximum suppression, resulting in a set of non-overlaping
segments $\widehat{x}_1,\ldots,\widehat{x}_n$. Each of these is fed to
the recognizer, producing
$\widehat{\letseq}_i=Rec(I_{\widehat{s}_i:\widehat{t}_i})$. Concatenating
these in the same way as in~\eqref{eq:lstar} gives us the full-video
predicted letter sequence $\widehat{\mathbf{L}}(\delta_f)$. We can now treat
$\mathbf{L}^\ast$ and $\widehat{\mathbf{L}}$ as two letter sequences,
and compute the transcription accuracy. Maximum sequence accuracy
(MSA) is defined as
\vspace{-.05in}
\begin{equation}
  \label{eq:msa}
    MSA=\displaystyle\max_{\delta_f}Acc(\mathbf{L}^\ast,\widehat{\mathbf{L}}(\delta_f)).
\end{equation}
Like AP@Acc, MSA depends on both the detector and the given recognizer
$Rec$.  By comparing the MSA for a given detector and for an ``oracle detector'' that produces the
ground-truth segments, we can obtain an indication of how far the detector output is from the ground-truth.

\section{Models for fingerspelling detection}

\kledit{Figure~\ref{fig:model_arch} sketches several baseline models and our proposed model, described below.} \kl{}

\subsection{Baseline models}

\textbf{Baseline 1: Frame-based detector}
This model classifies every frame as positive
(fingerspelling) or negative (non-fingerspelling), and is
trained using the per-frame cross-entropy loss, weighted to control
for class imbalance.  Each frame is passed through a convolutional network to extract visual features, which are then passed to a multi-layer bidirectional \kledit{long short-term memory RNN (LSTM)} to take temporal context into account, followed by a linear/sigmoid layer producing per-frame class posterior probabilities.

To convert the frame-level outputs to predicted segments, we first assign
hard binary frame labels by thresholding the posteriors by $\bar{p}$. Contiguous sequences of positive labels are taken to be the
predicted segments, with a segment score $f$ 
computed as the average posterior of the fingerspelling class for its
constituent frames.
We repeat this process for decreasing values of $\bar{p}$, producing a
pool of (possibly overlapping) segments. Finally, these are culled using 
thresholding on $f$ and
non-maximum suppression limiting overlap, producing a set of
$\widehat{x}$s as the final output.

\textbf{Baseline 2: Recognition-based detector}
Instead of classifying each frame into two classes, this baseline addresses the task as a sequence prediction problem, where the target is the
concatenation of
true letter sequences separated by $\emptyset$ indicating
non-fingerspelling segments, as
in~\eqref{eq:lstar}.
The frame-to-letter alignment in fingerspelling spans is usually
unknown during training, so we base the model on connectionist temporal classification (CTC)~\cite{graves2006ctc}: We generate frame-level label softmax posteriors over possible letters,
augmented by $\emptyset$ and a special $blank$ symbol, and train by
maximizing the marginal log probability of label sequences that
produce the true sequence under CTC's ``label collapsing function''
that removes duplicate frame labels and then $blank$s (we refer to
this as the CTC loss).

In this case we have a partial alignment of sequences and frame labels, since we know the
boundaries between $\emptyset$ and fingerspelling segments, and we 
use this information to explicitly compute a frame-level log-loss in
non-fingerspelling regions. This modification stabilizes training and improves performance.
At test time, we use the model's per-frame posterior
probability of fingerspelling, $1-p(\emptyset)$, and follow the same
process as in baseline 1 (frame-based) to convert the per-frame
probabilities to span predictions.

\textbf{Baseline 3: Region-based detector} 
This model directly predicts variable-length temporal segments that potentially contain fingerspelling, and is adapted from R-C3D~\cite{xu2017rc3d}, a 3D version of the Faster-RCNN~\cite{ren2015fasterrcnn}.
The model first applies a 2D ConvNet on each frame; additional 3D convolutional layers are applied on the whole feature tensor to capture temporal information. Unlike~\cite{xu2017rc3d}, we did not directly apply an off-the-shelf 3D ConvNet such as C3D~\cite{Tran2015LearningSF}, since its large stride may harm our ability to capture the delicate movements and often short sequences in fingerspelling.

A \kledit{region proposal network} \kl{} is applied \kledit{to the} feature tensor,
predicting shifts of potential fingerspelling segments with
respect to (temporal, 1D) anchors, and a binary label indicating
whether the predicted proposal contains fingerspelling. The
  detector is trained with a loss composed of two terms for (binary)
  classification and (positional) regression, $L_{det} = L_{cls} +
  L_{reg}$.
  See~\cite{xu2017rc3d} for further details.
At test time, we use greedy non-maximum suppression (NMS) on fingerspelling proposals to eliminate highly overlapping and low-confidence intervals.
Since fingerspelling detection is a binary classification task, we do not use a second stage classification subnet as in the original RC3D~\cite{xu2017rc3d}.

\subsection{Our \kledit{approach: A multi-task model}}

Our model
is based on the region-based detector, with the key difference being that fingerspelling recognition and
\kledit{pose estimation} are incorporated into training the model.

\textbf{Recognition loss} is computed by passing 
\bsedit{the fingerspelling segments} to a recognizer. 
Our intuition is that including the recognition task may help
the model learn richer features for fingerspelling, which may improve its ability to distinguish between fingerspelling and non-fingerspelling.
The recognizer here plays a role similar to the classification
subnet in RC3D. But since we don't assume that frame-letter alignment is
available at training time, we directly build a sub-network for
letter sequence prediction (the \kledit{orange ``REC''} in Figure~\ref{fig:model_arch}).

The recognition sub-network follows the attention-based model proposed for
fingerspelling recognition in~\cite{shi2019iccv}, using only the ground-truth segment for the recognition loss.
The recognition loss $L_{rec}$ is computed as the CTC loss summed over
the proposed regions predicted by the detector:
\vspace{-.05in}
\begin{equation}
 \label{eq:model_rec_ctc_loss}
 L_{rec} = \displaystyle\sum_{i=1}^nL_{ctc}(Rec(I_{s_i^\star:t_i^\star}), \letseq^\ast_i)
\end{equation}
where $n$ is the number of true fingerspelling segments.\bs{}

\textbf{Letter error rate loss} Though $L_{rec}$ should help
learn an improved image feature space, the performance of the detector does not impact the
recognizer directly since $L_{rec}$ uses only the ground-truth segments.
$L_{det}$ encourages the
model to output segments that are spatially close to the ground truth.
What's missing is the objective of making the detector work well
\emph{for the downstream recognition}. To this end we add a loss measuring the letter
error rate of a recognizer applied to proposals from the detector:
\begin{equation}
  \label{eq:model_rec_ler_loss}
  L_{ler}=-\sum_{i=1}^m p(\widehat{x}_i)Acc(\letseq^\ast_{k(i)},Rec(I_{\widehat{s}_i:\widehat{t}_i})),
\end{equation}
\kl{}
where as before, $k(i)$ is the index of the ground-truth segment matched (based
on IoU) to the
$i^\text{th}$ proposal $\widehat{x}_i=(s_i, t_i)$, $m$ is the number of proposals output by the detector, and $Acc$ is the \kledit{recognizer accuracy.}
The loss can be interpretated as the expectation of the negative
letter accuracy of the recognizer on the proposals given to it by
  the detector.
  Since $Acc$, which depends on
the edit distance, is non-differentiable, we approximate the gradient
of~\eqref{eq:model_rec_ler_loss}
\kledit{as in} 
REINFORCE~\cite{Williams2004SimpleSG}:
\begin{equation}
  \label{eq:model_rec_ler_loss_grad}
  \nabla L_{ler}\approx
  -\sum_{i=1}^Mp(\widehat{x}_i)Acc(\letseq^\ast_{k(i)},Rec(I_{\widehat{s}_i:\widehat{t}_i})\nabla\log p(\widehat{x}_i),
\end{equation}
where the sum is over the $M$ highest scoring proposals, and
$p(\widehat{x}_i)$ is the normalized score $f_i/\sum_{i=1}^Mf_i$.

\textbf{Pose estimation loss} Since sign language is
to a large extent based on body articulation,
it is natural to consider
incorporating pose
into the model.
However, we can not assume access to ground-truth \kledit{pose}
even in the training data.
Instead, we can rely on general-purpose human pose
estimation models, such as OpenPose~\cite{cao2019openpose}, trained on large sets of annotated images.

Prior work on sign language
has used pose estimates
extracted from an external model as input to models for sign
language-related tasks~\cite{Joze2019MSASLAL,Moryossef2020sld,Parelli2020slpose,li2020word,Charles2013AutomaticAE}. On the controlled studio data used in that work,
the quality of extracted keypoints is reliable. On more challenging
real-world data like ChicagoFSWild/ChicagoFSWild+~\cite{shi2019iccv}, we find the
detected keypoints from OpenPose to be much less reliable, 
presumably in part due to
the widespread motion blur in the data (see Figure~\ref{fig:pose_samples} for examples).  \kledit{Indeed, as we show in our experiments, using automatically extracted poses as input to the model does not significantly improve performance.}

Instead of relying on estimated pose at test time, we treat
the estimated pose as a source of additional supervision for our model
at training time. We use keypoints from OpenPose as pseudo-labels to
help distill knowledge from the pre-trained pose model into
our detection model. As pose is \kledit{used only} as an auxiliary task, the
quality of those pseudo-labels has less impact on detection
performance than does using pose as input at test time.

The pose estimation sub-network takes the feature maps extracted by the model (shared with
the detector and recognizer) and applies several transposed
convolutional layers to increase spatial resolution, producing for
frame $I_t$ a set of heat
maps $\mathbf{b}_{t,1},\ldots,\mathbf{b}_{t,P}$ for $P$ keypoints in the OpenPose
model. We also use OpenPose to extract keypoints from $I_t$; each
estimated keypoint $p$ is accompanied by a confidence score
$\sigma_{t,p}$. We convert these estimates to pseudo-ground truth heatmaps
$\mathbf{b}^\ast_{t,1},\ldots,\mathbf{b}^\ast_{t,P}$.  The pose loss is the
per-pixel Euclidean distance between the predicted and pseudo-ground truth maps:
\begin{equation}
  \label{eq:model_pose_loss}
  L_{pose}=\displaystyle\sum_{t=1}^{T}\displaystyle\sum_{p=1}^{P}\|\v
  b_{t,p} -\v b^\ast_{t,p}\|^2\cdot 1_{\sigma_{t,p}>\tau},
\end{equation}
where the threshold on the OpenPose confidence is used to ignore
low-confidence pseudo-labels.

Putting it all together, the loss for training our model is
\begin{equation}
  \label{eq:model_total_loss}
  L=L_{det} + \lambda_{rec} L_{rec} + \lambda_{ler} L_{ler} + \lambda_{pose} L_{pose},
\end{equation}
with tuned weights $\lambda_{ler}$, $\lambda_{rec}$,
$\lambda_{pose}$ controlling the relative importance of different loss components.\bs{} %

\textbf{Second stage refinement}
Since we expect reasoning about
fine-grained motion and handshape differences to be helpful,
we would like to use high-resolution input
images. However, since the input video clip covers hundreds of
frames, the images need to be downsampled to fit into the memory
of a typical GPU.

To mitigate the issue of low resolution in local regions, we
\kledit{``zoom in'' on} the original image frames using the attention map produced by the reocognizer sub-network.  This idea is based on the iterative attention approach of~\cite{shi2019iccv}.  As large attention values indicate higher importance \kledit{of} the corresponding region for fingerspelling, cropping the ROI surrounding \kledit{a location with high attention values} helps increase the resolution of that part while avoiding other irrelevant areas.  To achieve this, we first complete training of the whole model described above. Suppose the image sequence of original resolution is $I^o_{1:N}$, the lower-resolution frame sequence used in the first round of training is $I^g_{1:N}$, and the trained model is $\mathcal{H}$. We run inference on $I^g_{1:N}$ with the recognition sub-network $\mathcal{H}_{rec}$ to produce a sequence of attention maps $\mathbf{A}_{1:N}$. We use  $\mathbf{A}_{1:N}$ to crop  $I^o_{1:N}$ into a sequence of local ROIs  $I^l_{1:N}$. Specifically, at timestep $n$, we put a bounding box $b_n$ of size $R|I_n|$ centered on the peak of attention map $\v A_n$, where $R$ is the zooming factor. We average \kledit{the} bounding boxes of \kledit{the} $2a+1$ frames centered on the \kledit{$n^\textrm{th}$} frame to produce a smoother cropping ``tube'' $b_{1:N}^s$:
\begin{equation}
  \label{eq:model_second_bbox}
  b^s_n=\frac{1}{2a+1}\displaystyle\sum_{i=-a}^{a}(b_{n+i})
\end{equation}
The local ROIs $I^{l}_{1:N}$ are cropped from $I^o_{1:N}$ with $b_{1:N}^s$.

Finally, we perform a second stage of training with both
$I^{g}_{1:N}$ and $I^{l}_{1:N}$
as input. At each timestep $n$, the backbone conv layers are applied
on $I^{g}_n$ and $I^{l}_n$ to obtain global and
local feature maps, respectively. The two feature maps
are concatenated for
detection and recognition. The pseudo-ground truth pose maps are estimated on $ I^{(g)}_t$ and $I^{(l)}_t$ separately. The overall loss for the second-stage training is
\begin{equation}
  \label{eq:model_second_loss}
  \begin{split}
    L^{\text{final}}\,=&\,L_{det} + \lambda_{rec} L_{rec} + \lambda_{ler} L_{ler} \\
    &+\,\lambda_{pose} (L^{(g)}_{pose} + L^{(l)}_{pose}) \\
  \end{split}
\end{equation}
The key difference between our approach and the iterative attention of~\cite{shi2019iccv} is that we do not drop the input images, but rather use the newly generated ROIs as extra input. In fingerspelling detection, both global context (e.g., upper body position) and local details (e.g., handshape) are important.

\section{Experiments}
\label{sec:exp}
\subsection{Setup}

We conduct experiments on ChicagoFSWild~\cite{shi2018slt} and
ChicagoFSWild+~\cite{shi2019iccv},
two large-scale ASL fingerspelling datasets collected in the wild.
Though the datasets were introduced for fingerspelling recognition (with the boundaries given), the URLs of the raw ASL videos and the fingerspelling start/end timestamps are provided.
We split each video clip into 300-frame chunks \kledit{($\sim$12s)} with a 75-frame overlap between chunks.
The longest fingerspelling sequence in the data is 290 frames long.
We use the same training/dev/test data split as in the
original datasets \kledit{(see additional data statistics in the supplementary material)}.
The image frames are
center-cropped and resized to $108\times 108$.

We
take the convolutional layers from VGG-19~\cite{Simonyan2015VeryDC}
pre-trained on ImageNet~\cite{imagenet_cvpr09} as our backbone
network, and fine-tune the weights during training. For baselines 1 and 2, an average pooling layer is applied on
the feature map, giving a 512-dimensional vector for each frame, which
is fed into a one-layer Bi-LSTM with 512 hidden units. In baseline 3
and our model, the feature map is further passed through a 3D conv +
maxpooling layer (with temporal stride 8). In \kledit{the region proposal network}, \kledit{the lengths of the anchors are fixed at 12 values ranging from 8 to 320,}
which are chosen according to the typical lengths of
fingerspelling sequences in the data. The IoU thresholds for
positive/negative anchors are respectively 0.7/0.3. The predicted
segments are refined with NMS at a threshold of 0.7. The magnitude of
optical flow is used as a prior attention map \kledit{as in~\cite{shi2019iccv}}.
\bsedit{We use the same recognizer ($Rec$) for~\eqref{eq:model_rec_ctc_loss} and ~\eqref{eq:model_rec_ler_loss}.}
The pose sub-net is composed of one transposed convolutional layer with
stride 2. We use OpenPose~\cite{cao2019openpose} to extract 15 body
keypoints and $2\times 21$ hand keypoints (both hands) as pseudo pose
labels. Keypoints with confidence below $\tau=0.5$ are dropped. For
\kledit{second-stage refinement,} the moving averaging of bounding boxes
in~\eqref{eq:model_second_bbox} uses 11 frames ($a=5$) \bsedit{and the frame sequence is downsampled by a factor of 2 to save memory.} \kledit{The loss weights ($\lambda$s) are tuned on the dev set.}

To evaluate with AP@Acc and MSA, we train a reference \kledit{recognizer}
with the same architecture as the recognition-based detector on
the fingerspelling training data.
\bsedit{The recognizer achieves an accuracy (\%) of 44.0/62.2 on
  ground-truth fingerspelling segments on
  ChicagoFSWild/ChicagoFSWild+.} \kledit{The accuracy is
  slightly worse than that of~\cite{shi2019iccv} because we used a
  simpler recognizer.} \bsedit{In particular, we skipped the iterative
  training in~\cite{shi2019iccv},
  used lower-resolution input, and did not use a language model.}
We consider $\delta_{IoU}\in\{0.1, 0.3, 0.5\}$ and $\delta_{acc}\in\{0,0.2,0.4\}$; $\delta_{IoU}$ is fixed at 0 for AP@Acc.

\subsection{Results}

\begin{table}[h]
  \caption{\label{tab:results}Model comparison on the ChicagoFSWild and
    ChicagoFSWild+ \bsedit{test sets}.  BMN refers to boundary matching network~\cite{lin2019bmn}.  The right column (GT) shows results when the
    ``detections'' are given by ground-truth fingerspelling segments.} 
  \setlength{\tabcolsep}{2pt}
  \small
  \begin{tabular}{cc|ccccc|c}\hline
    \multicolumn{2}{c|}{ChicagoFSWild} & Base 1 & Base 2 & Base 3 & BMN & Ours & GT \\ \hline
   \multirow{3}{*}{\shortstack{AP@\\IoU}} & AP@0.1 & .121 & .310 & .447 &.442 & \textbf{.495}  & 1.00  \\
   & AP@0.3 & .028 & .178 & .406 & .396 & \textbf{.453} & 1.00 \\
   & AP@0.5 & .012 & .087 & .318 & .284 & \textbf{.344} & 1.00 \\ \hline
   \multirow{3}{*}{\shortstack{AP@\\Acc}} & AP@0.0 & .062 & .158 & .216 & .209 & \textbf{.249}  & .452  \\
   & AP@0.2 & .028 & .106 & .161 & .157 & \textbf{.181} & .349 \\
   & AP@0.4 & .006 & .034 & .069 & .070 & \textbf{.081} & .191 \\ \hline
   \multicolumn{2}{c|}{MSA} & .231 & .256 & .320 & .307 & \textbf{.341} & .452 \\ \hline\hline
       \multicolumn{2}{c|}{(b). ChicagoFSWild+} & Base 1 & Base 2 & Base 3 & BMN & Ours & GT \\ \hline
   \multirow{3}{*}{\shortstack{AP@\\IoU}} & AP@0.1 & .278 & .443 & .560 & .580 & \textbf{.590}  & 1.00  \\
   & AP@0.3 & .082 & .366 & .525 & .549 & \textbf{.562} & 1.00 \\
   & AP@0.5 & .025 & .247 & .420 & .437 & \textbf{.448} & 1.00 \\ \hline
   \multirow{3}{*}{\shortstack{AP@\\Acc}} & AP@0.0 & .211 & .323 & .426 & .433 & \textbf{.450}  & .744  \\
   & AP@0.2 & .093 & .265 & .396 & .401 & \textbf{.415} & .700 \\
   & AP@0.4 & .029 & .152 & .264 & .260 & \textbf{.277} & .505 \\ \hline
   \multicolumn{2}{c|}{MSA} & .267 & .390 & .477 & .470 & \textbf{.503} & .630 \\ \hline
  \end{tabular}
  \vspace{-0.1in}
\end{table}

\textbf{Main results} Table~\ref{tab:results} compares models using the three proposed metrics. The values of AP@Acc and MSA depend on the reference recognizer. For ease of comparison, we also show the oracle results when the same recognizer is given the ground-truth fingerspelling segments.
Overall, the relative model performance is consistent across metrics.  Methods that combine detection and recognition outperform those that do purely detection (baseline 2 vs.~1, our model vs.~baseline 3).
In addition, region-based methods (baseline 3 and our model) outperform frame-based methods (baseline 1 \& 2), whose segment predictions lack smoothness.  We can see this also by examining the frame-level performance of the three baselines.\footnote{For the region-based model (baseline 3) we take the maximum probability of all proposals containing a given frame as that frame's probability.}  Their frame-level average precisions are 0.522 (baseline 1), 0.534
(baseline 2), and 0.588 (baseline 3), which are much closer than the
segment-based metrics, \kledit{showing how frame-level metrics can obscure important differences between models.}
\bsedit{More details on precision and recall can be found in the supplementary material.}
\kledit{The trends in results are similar on the two datasets. For the remaining analysis we report results on the ChicagoFSWild dev set.}

In addtion to baselines adapted from related tasks,
  Table~\ref{tab:results} includes a comparison to a recent method
  developed for action detection: the boundary matching
  network~\cite{lin2019bmn}. We also compared to the multi-stage
  temporal convolutional network~\cite{farha2019mstcn} (details in the
  supplementary). Neither
  approach is better than ours on the fingerspelling benchmark.
  \bs{} \kl{}

\begin{figure}[!tb]
  \begin{minipage}[c]{.7\linewidth}
  \includegraphics[width=0.99\linewidth]{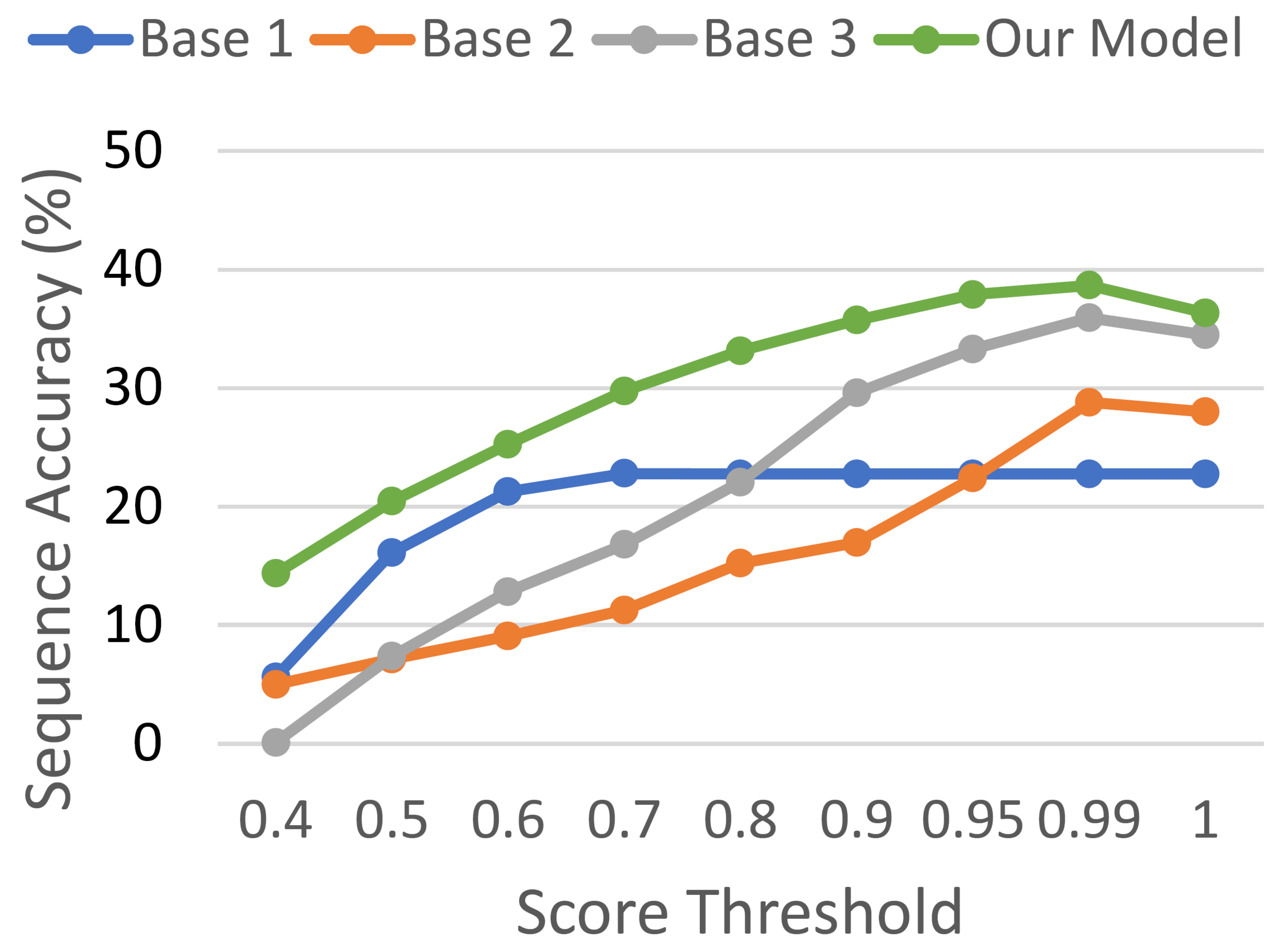}    
\end{minipage}%
\begin{minipage}[c]{.3\linewidth}
  \caption{\label{fig:sa_thr}Dependence of sequence accuracy on score
    threshold $\delta_f$. \kl{}}  
\end{minipage}
\vspace{-0.2in}
\end{figure}

\textbf{Analysis of evaluation metrics} \kledit{The AP@Acc results
are largely invariant to $\delta_{IoU}$ (see supplementary material)
and we report} results for $\delta_{IoU}=0$ (i.e., matching ground truth segment to detections
based on accuracy, subject to non-zero overlap). 
We also examine the relationship between sequence accuracy and the
score threshold of each model (Figure~\ref{fig:sa_thr}). Our model
achieves higher sequence accuracy across all thresholds. The threshold
producing the best accuracy varies for each model.
The average IoUs of the four models for the optimal threshold
$\delta_f$ are 0.096, 0.270, 0.485, and 0.524 respectively.

\begin{figure}[b]
\vspace{-0.2in}
  \includegraphics[width=\linewidth]{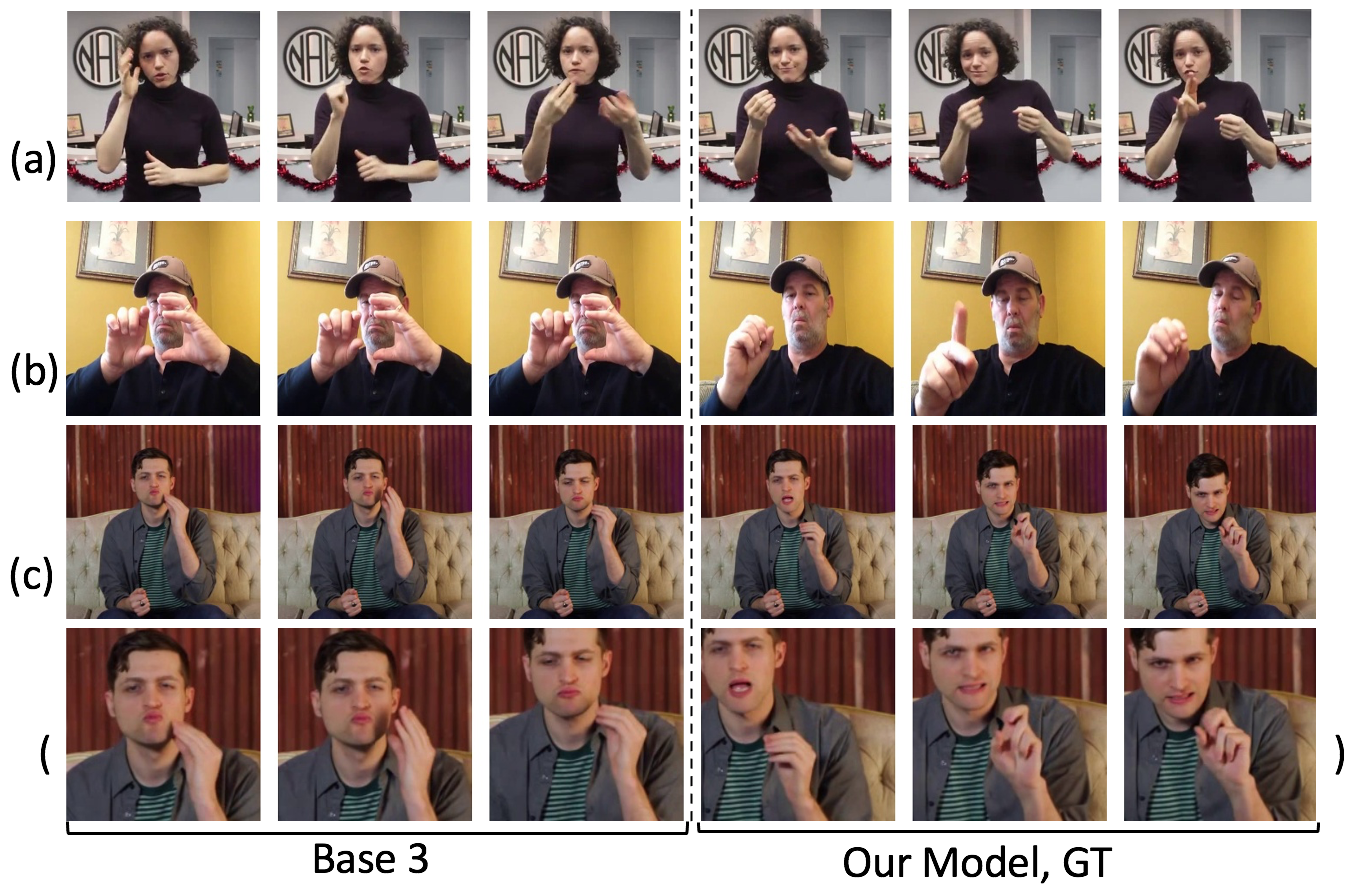}
  \caption{\label{fig:quals}Example segments detected by different
    models. Bottom row: ROIs used in \kledit{second-stage refinement}. GT: Ground-truth segment. \bsedit{The sequence is downsampled.}}
\end{figure}

\textbf{Ablation study} Our model reduces to baseline 3 when all loss terms except the detection loss are removed.  Table~\ref{tab:analysis_loss} shows how model performance improves as more tasks are added. The gain due to the recognition loss alone is smaller than for the frame-based models (base 1 vs. base 2). The recognition sub-network contributes more through the LER loss, which communicates the recognition error to the detector directly. The second-stage refinement does not always boost AP@IoU; the gain is more consistent in the accuracy-based metrics (AP@Acc, MSA).
The zoom-in effect of the second-stage refinement increases the resolution of the hand region and improves recognition,
though the IoU
may remain similar.
The downsampling in the second-stage refinement also leads to some positioning error, reflected in the slight drop in AP at IoU=0.5, though a minor temporal shift does not always hurt recognition.

\begin{table}[t]
  \caption{\label{tab:analysis_loss}Impact of adding loss components
    to our model training.  Results are on \bsedit{the ChicagoFSWild dev set}. \kl{}}
  \setlength{\tabcolsep}{3pt}
  \begin{tabular}{cc|c|c|c|c|c}\hline
    & & Base 3 & + rec & +LER & +pose & +2nd stage \\ \hline
    \multirow{3}{*}{\shortstack{AP@\\IoU}} & AP@0.1 & .496 & .505 & .531 & .539  & .551  \\
    & AP@0.3 & .466 & .479 & .498 & .500 & .505 \\
    & AP@0.5 & .354 & .362 & .392 & .399 & .393 \\ \hline
    \multirow{3}{*}{\shortstack{AP@\\Acc}} & AP@0.0 & .285 & .290  & .299  & .302  & .313  \\
    & AP@0.2 & .222 & .231 & .245  & .255 & .267 \\
    & AP@0.4 & .094 & .097 & .103  & .107 & .112 \\ \hline
    \multicolumn{2}{c|}{MSA} & .359 & .365  & .378 & .383 & .386 \\ \hline
  \end{tabular}
  \vspace{-0.1in}
\end{table}

{\bf Examples} Figure~\ref{fig:quals} shows examples in which our
model correctly detects fingerspelling segments while baseline 3
fails. Fingerspelling can include handshapes that are visually similar
to non-fingerspelled signs (Figure~\ref{fig:quals}a). Fingerspelling
recognition, as an auxiliary task, may improve detection by helping
the model distinguish among fine-grained handshapes. \kledit{The signer's pose}
may provide additional information for detecting fingerspelling (Figure~\ref{fig:quals}b). Figure~\ref{fig:quals}c shows an example where the signing hand is a small portion of the whole image; baseline 3 likely fails due to the low resolution of the signing hand. The second-stage refinement enables the model to access a higher-resolution ROI, leading to a correct detection.

\begin{figure}[b]
  \centering
  \includegraphics[width=\linewidth]{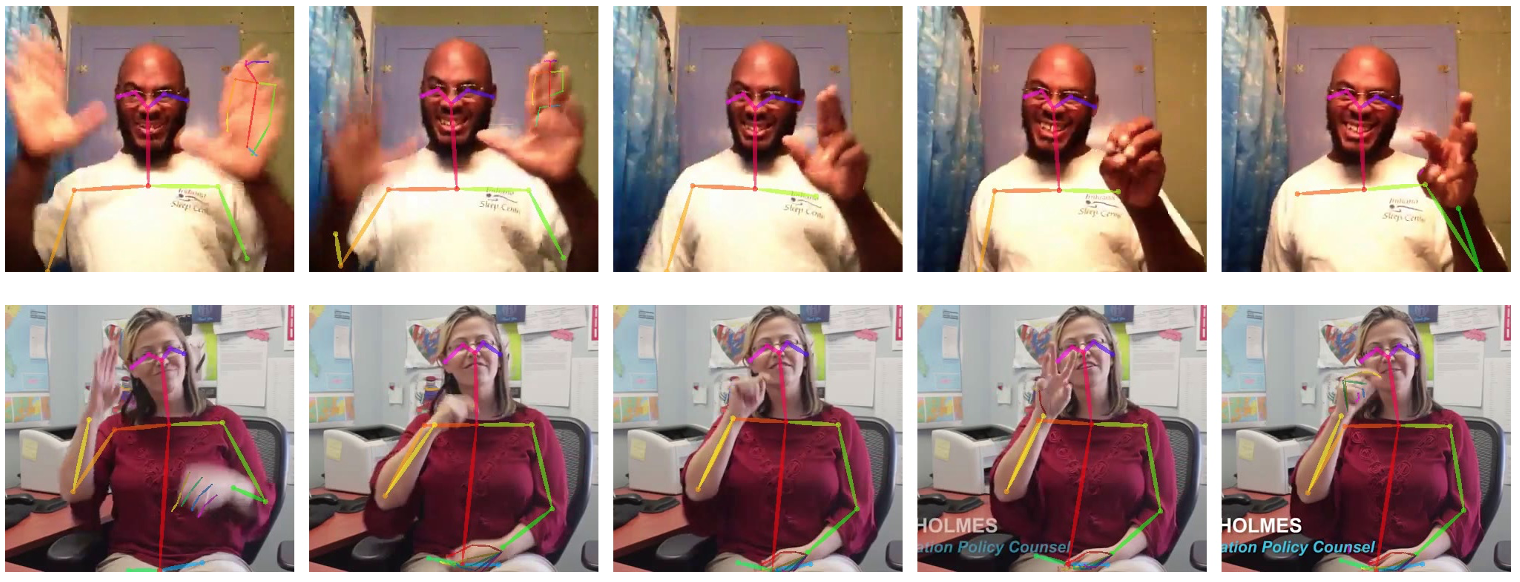}
    \caption{\label{fig:pose_samples} \kledit{Pose estimation failure cases.}}
\end{figure}

\begin{figure}[b]
  \begin{minipage}{0.04\linewidth}
    (a),
  \end{minipage}
  \begin{minipage}{0.96\linewidth}
    \includegraphics[width=\linewidth]{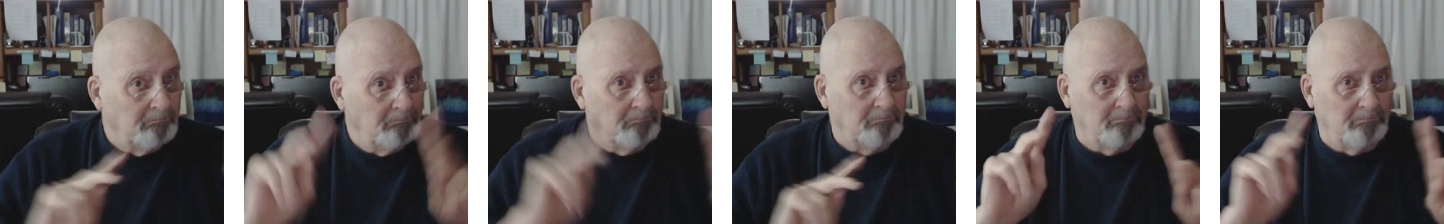}
  \end{minipage}\\ 
  \begin{minipage}{0.04\linewidth}
    (b),
  \end{minipage}
  \begin{minipage}{0.96\linewidth}
    \includegraphics[width=\linewidth]{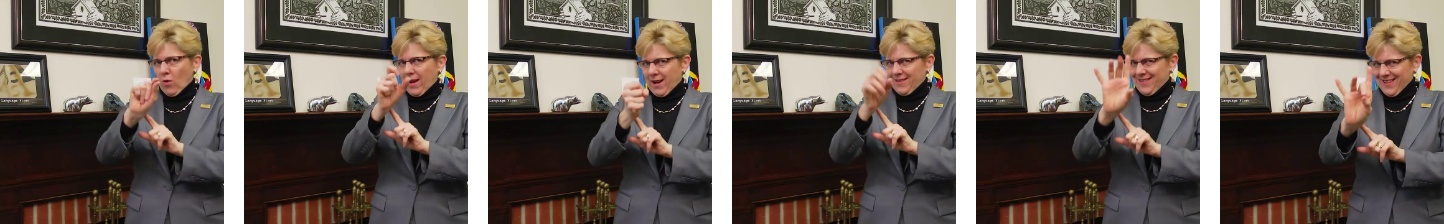}
  \end{minipage}\\ 
  \begin{minipage}{0.04\linewidth}
    (c),
  \end{minipage}
  \begin{minipage}{0.96\linewidth}
    \includegraphics[width=\linewidth]{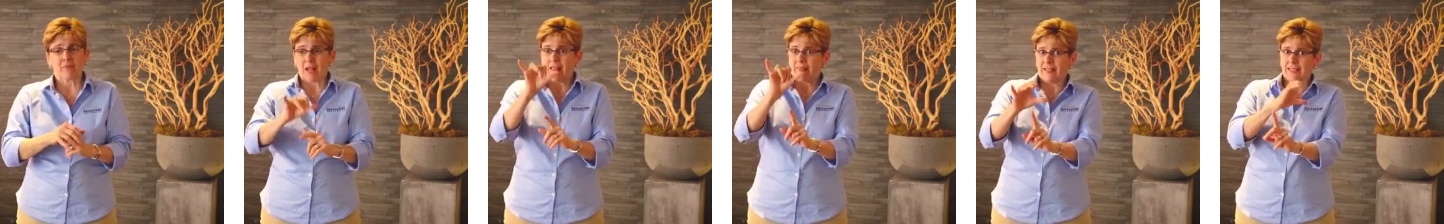}
  \end{minipage}
  \caption{\label{fig:fps}False positive detections. The glosses in (a), (b), (c) are respectively ``PEOPLE'', ``2018'', ``THAT-[C..\}''. \kl{}}
\end{figure}

\textbf{Error analysis} 
Qualitatively, we notice three common sources of false positives: (a) near-face sign, (b) numbers, and (c) fingerspelling handshape used in regular signing \kledit{(see Figure~\ref{fig:fps})}. Such errors can potentially be reduced by incorporating linguistic knowledge in the detector, which is left as future work. \kledit{We also observe that short fingerpelling segments are harder to detect. See the supplementary material for the performance breakdown according to length.}

\textbf{\kledit{Other input modalities}} \kl{}\bs{} Our model is trained on RGB images. Motion and pose are two common modalities used in sign language tasks~\cite{Joze2019MSASLAL,Moryossef2020sld,yanovich2016detection}, so it is natural to ask whether they would be helpful instead of or in addition to RGB input.
We use magnitude of optical flow~\cite{Farnebck2003TwoFrameME} as the motion image and the pose heatmap from OpenPose as the pose image.
\kledit{Raw RGB frames as input outperform the other two modalities, although each of them can be slightly helpful when concatenated with the RGB input (see Table~\ref{tab:modalities}).}
The pose image is less consistently helpful, likely because pose estimation is very challenging on our data and OpenPose often fails to detect the keypoints especially in the signing hand (see Figure~\ref{fig:pose_samples}).  However, treating pose estimation as a secondary task, as is done in our model, successfully ``distills'' the \kledit{pose model's} knowledge
and outperforms the use of additional modalities as input. \kledit{We note that using all three modalities can boost performance further. Using both RGB and motion images as input while jointly estimating pose, the detector achieves .523/.495/.367 for AP@IoU(0.1/0.3/0.5), improving over the best model in Table~\ref{tab:modalities}.  However, optimizing performance with multiple modalities is not our main focus, and we leave further study of this direction to future work.}

\begin{table}[t]
    \caption{\label{tab:modalities}Comparison among modalities for the region-based detector (baseline 3) \bsedit{on the ChicagoFSWild dev set}. RGB+Pose(in): both RGB and pose image used as input. RGB+Opt(in): both RGB and optical flow used as input. RGB+Pose(out): detector trained jointly with pose estimation \kledit{(our model)}.} 
  \centering
  \setlength{\tabcolsep}{1.5pt}
  \small
  \begin{tabular}{c|cccccc}\hline
    AP@IoU & RGB & Pose & Opt & \shortstack{RGB+Opt\\ (in)} & \shortstack{RGB+Pose\\ (in)} & \shortstack{RGB+Pose\\ (out)} \\ \hline
    AP@0.1 & .496 & .368 & .476 & .503 & .501 & \textbf{.505} \\
    AP@0.3 & .466 & .332 & .438 & .472 & .470 & \textbf{.478} \\
    AP@0.5 & .354 & .237 & .315 & .357 & .346 & \textbf{.366} \\ \hline
  \end{tabular}
  \vspace{-0.1in}
\end{table}

\section{Conclusion}

We study the task of sign language fingerspelling detection, motivated
by the 
goal of developing it as an upstream component in an automatic
recognition system. We propose a benchmark, based on extending a
previously released fingerspelling recognition dataset, and establish a
suite of metrics to evaluate detection models both on their own merits
and on their contribution to downstream recognition. We investigate
approaches adapted from frame classification, fingerspelling
recognition and action detection, and demonstrate that a novel,
multi-task model achieves
the best results across metrics. Beyond
standard detection loss, this model incorporates losses derived from
recognition and pose estimation; we show that each of these
contributes to the superior performance of the model.

Our results provide, for the first time, a practical recipe for fully
automatic detection and recognition of fingerspelling in real-world
ASL media. While our focus has been on fingerspelling, we expect that
the techniques, both in training and in evaluation of the methods,
will be helpful in the more general sign language detection and
recognition domain, which remains our goal for future work.


{\small
\bibliographystyle{ieee_fullname}
\bibliography{egbib}
}

\clearpage

\appendix
\bigskip
{\noindent \large \bf {Supplementary Material}}\\

\section{Statistics of the datasets\bs{}}

\begin{table}[htp]
  \centering
  \setlength{\tabcolsep}{3pt}
  \caption{\label{tab:data_stats}Numbers of 300-frame raw ASL clips in the ChicagoFSWild and ChicagoFSWild+ data subsets. \kledit{The number of fingerspelling segments in each subset is given in parentheses.}}
  \begin{tabular}{l|ccc}\hline
    & train & dev & test \\ \hline
    ChicagoFSWild & 3539 (6927) & 691 (1246) & 613 (1102) \\ \hline
    ChicagoFSWild+ & 13011 (44861) & 867 (2790) & 885 (1531) \\ \hline
  \end{tabular}
\end{table}

\begin{figure}[ht]
 \centering
 \caption{\label{fig:fs-length-dist}Distribution of length of fingerspelling segments.}
 \includegraphics[width=\linewidth]{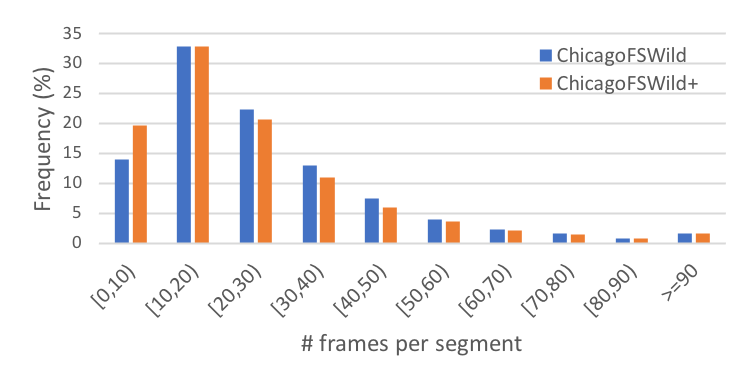}
\end{figure}

\kledit{Table~\ref{tab:data_stats} provides the numbers of clips and of fingerspelling segments in the datasets used in our work.  Note that the number of fingerspelling segments is not exactly same as in~\cite{shi2019iccv,shi2018slt} due to the 75-frame overlap when we split raw video into 300-frame clips. On average there are 1.9/1.8 fingerspelling segments per clip for ChicagoFSWild/ChicagoFSWild+. The distributions of durations are shown in Figure~\ref{fig:fs-length-dist}.}

\section{Precision-recall \kledit{curves for frame classification}}
\begin{figure}[htb]
 \centering
 \caption{\label{fig:pr_curve_frame}Precision-recall \kledit{curves for frame classification of three baselines and} our approach.}
 \includegraphics[width=\linewidth]{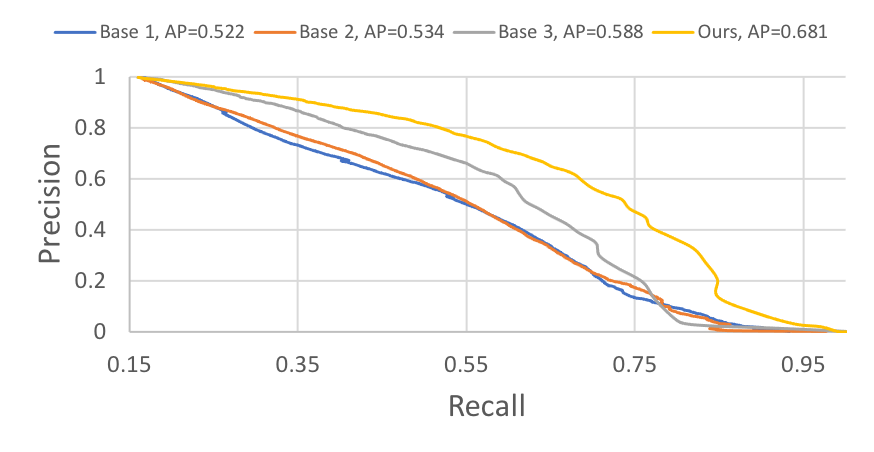}
\end{figure}

Figure~\ref{fig:pr_curve_frame} shows the precision-recall \kledit{curves for frame classification of the three baseline models and our proposed model. On the one hand, our approach dominates the others in terms of these frame-level metrics as well. In addition, the differences in frame-level performance among the three baselines are much smaller than the differences in sequence-level performance reported in the main text.} 

\section{Comparison with \kledit{state-of-the-art models for} related tasks\bs{}}
\kledit{In addition to the baseline models, we compare against two additional approaches---boundary matching network (BMN)~\cite{lin2019bmn} and multi-stage temporal convolutional network (MS-TCN)~\cite{farha2019mstcn}---on our task of fingerspelling detection. Those two methods are state-of-the-art on temporal action proposal generation in ActivityNet1.3~\cite{caba2015activitynet} and sign language segmentation~\cite{Renz2021signsegmentation}. The implementations are based on~\cite{bmn-code,mstcn-code}. For fair comparison, we use the same backbone network as in the other methods. We use the same network architecture for the individual submodules of the two models and tune hyperparameters on our datasets. As MS-TCN does frame classification in principle, we follow the same post-processing steps as in baseline 1 and 2 to convert frame probabilities into sequence predictions for evaluation.}

\kledit{As is shown in Table~\ref{tab:res-bmn-mstcn}, these two approaches do not outperform our approach. Comparing BMN and our baseline 3, we notice that the size of the training set has a large impact. The more complex modeling choices in BMN, which searches over a wider range of proposals, leads to better performance mostly when using the larger training set of ChicagoFSWild+. The discrepancy in performance of these two models as measured by different metrics (e.g., AP@IoU vs.~AP@Acc) also shows that a model with lower localization error does not always enable more accurate downstream recognition. The MS-TCN model is generally better than other frame-based approaches (baseline 1, 2) but remains inferior to region-based approaches including baseline 3 and ours. Our post-processing steps lead to inconsistency between the training objective and evaluation. Similarly in~\cite{Renz2021signsegmentation}, it is noted that the model sometimes over-segments fingerspelled words.}

\begin{table}[h]
    \caption{ Performance of BMN and MS-TCN on fingerspelling detection using our evaluation metrics on the (a) ChicagoFSWild and (b) ChicagoFSWild+ test sets. \bs{}\kl{}}
    \label{tab:res-bmn-mstcn}
    \centering
    \footnotesize
    \setlength{\tabcolsep}{1pt}
    \begin{tabular}{cc|ccc|ccc|c}\hline
    \multirow{2}{*}{} & \multirow{2}{*}{} &
         \multicolumn{3}{c|}{AP@IoU} & \multicolumn{3}{c|}{AP@Acc} & \multirow{2}{*}{MSA}\\ 
         & & AP@0.1 & AP@0.3 & AP@0.5 & AP@0.0 & AP@0.2 & AP@0.4 & \\ \hline
         \multirow{2}{*}{(a)}& BMN & .442 & .394 & .284 & .209 & .157 & .070 & .307 \\ \cline{2-9}
& MS-TCN & .282 & .177 & .095 & .141 & .093 & .036 & .319 \\ \hline
\multirow{2}{*}{(b)} & BMN & .580 & .549 & .437 & .433 & .401 & .260 & .470 \\ \cline{2-9}
& MS-TCN & .429 & .345 & .179 & .350 & .299 & .147 & .414 \\ \hline
    \end{tabular}
\end{table}

\section{Analysis of AP@Acc}
\begin{figure}[h]
  \centering
  \caption{\label{fig:ap_heatmap}AP@Acc with different IoU thresholds \bsedit{on ChicagoFSWild dev set}. Left: baseline 3. Right: our model.}
  \includegraphics[width=\linewidth]{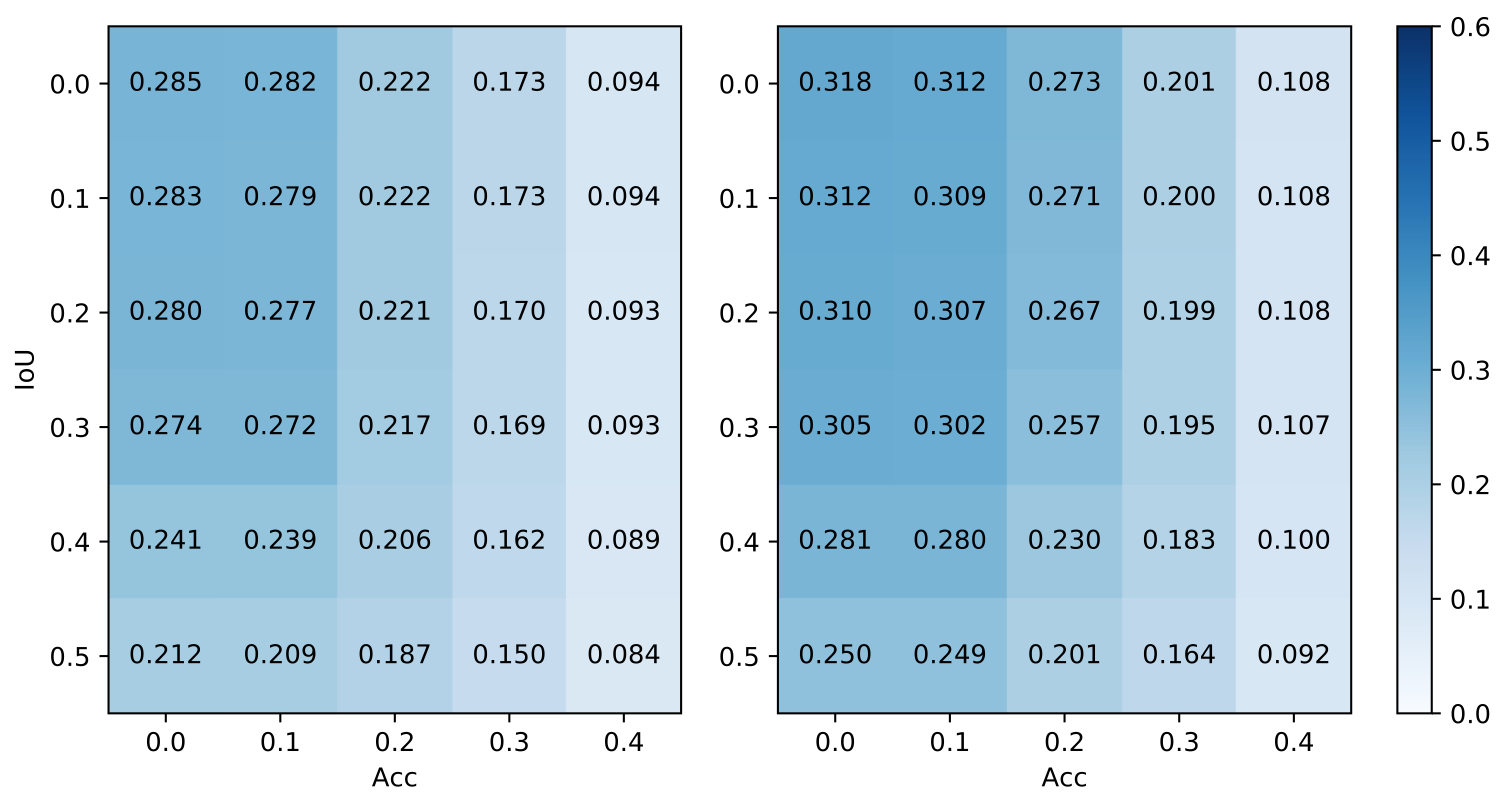}
\end{figure}

Figure~\ref{fig:ap_heatmap} shows how varying $\delta_{IoU}$ and $\delta_{acc}$ impacts the value of \kledit{AP@Acc}.
The accuracy threshold $\delta_{acc}$ has a much larger impact on AP than does $\delta_{IoU}$. This is primarily because a \kledit{large} overlap between predicted and ground-truth segments is often necessary in order to achieve high accuracy. Therefore, we set the default value of $\delta_{IoU}$ to 0.

\section{Histogram of IoU}
\begin{figure}[h]
 \centering
 \caption{Histogram of IoU at peak thresholds.}
 \includegraphics[width=1\linewidth]{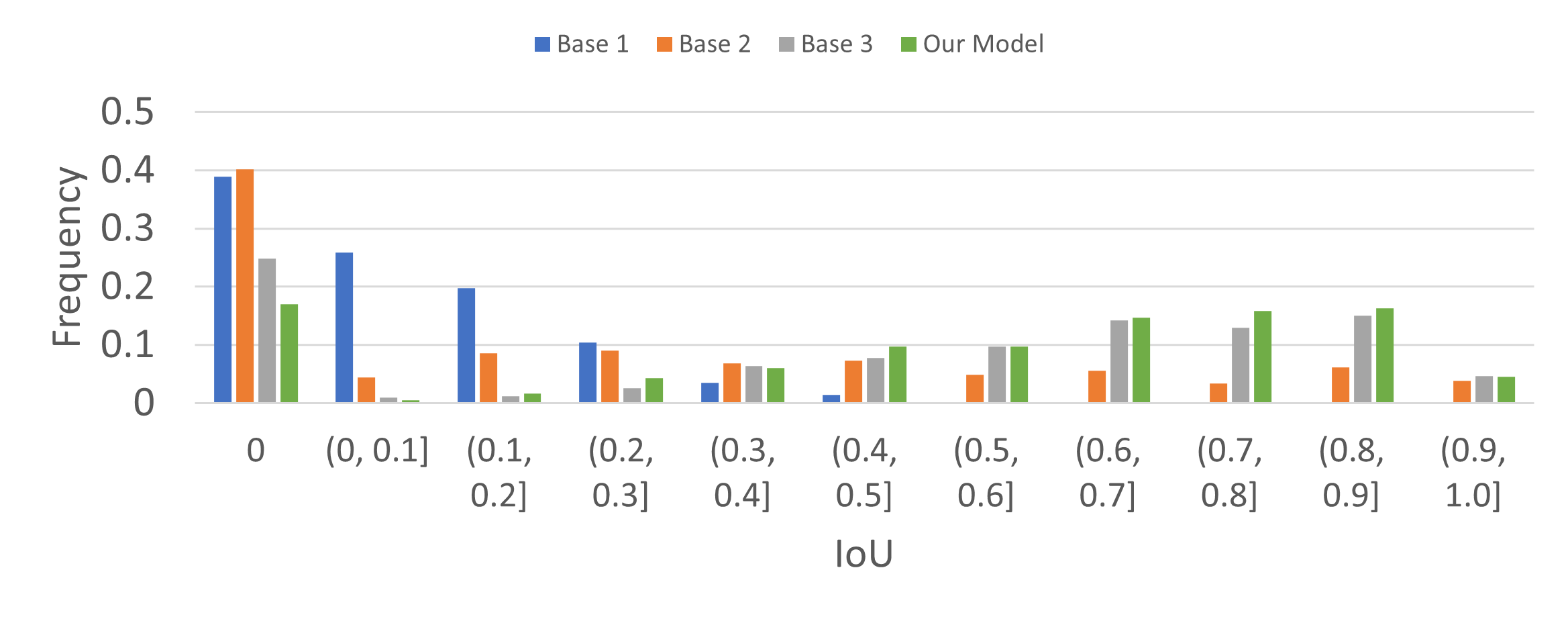}
 \label{fig:iou_histo}
\end{figure}

\kledit{Figure~\ref{fig:iou_histo} shows histograms of IoU of predicted segments with respect to the ground truth at peak thresholds used in the MSA computation.
Our model has overall higher IoU than the three baselines.
The average IoUs of the three baselines and our model for the optimal (peak) threshold
$\delta_f$ are 0.096, 0.270, 0.485, and 0.524 respectively.}
\bsedit{The average IoUs of baseline 3 and our model} suggest that for AP@IoU, AP@0.5 is more meaningful to compare in terms of recognition performance for \bsedit{those two models}.\kl{}\bs{}

\section{Performance breakdown over durations\bs{}}

\begin{figure}[htb]
  \centering
  \caption{\label{fig:fp_fn_distribution}Distribution of lengths of false positives and false negatives.}
  \includegraphics[width=\linewidth]{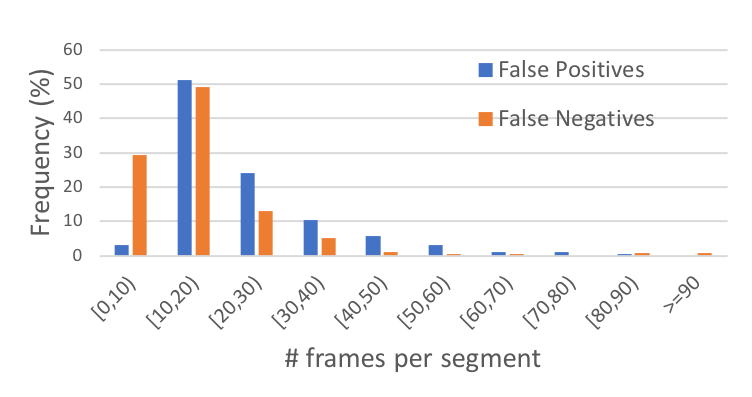}
\end{figure}

\kledit{We separate raw video clips into three categories based on the duration of the fingerspellng segments: short ($<$20 frames), medium (20-80 frames), and long ($\geq$80 frames). This division is based on the statistics of the dataset. The performance of our model for the three categories is shown in Table~\ref{tab:perf_dur}, and can be compared to the overall performance in \kledit{Table 1} of the paper.\kl{} Shorter fingerspelling segments are harder to spot within regular signing. The typical fingerspelling pattern (relatively static arm and fast finger motion) is less obvious in short segments. In addition, Figure~\ref{fig:fp_fn_distribution} shows the length distribution of false positive and false negative detections from our model.} The length distribution of false positives roughly matches that of ground-truth segments in the dataset. 

 \begin{table}[htp]
    \caption{ Performance on segments of different durations.\bs{}}
    \label{tab:perf_dur}
    \centering
    \footnotesize
    \setlength{\tabcolsep}{1.5pt}
    \begin{tabular}{c|c|c|c|c|c|c|c}\hline
    \multirow{2}{*}{} &
         \multicolumn{3}{c|}{AP@IoU} & \multicolumn{3}{c|}{AP@Acc} & \multirow{2}{*}{MSA}\\ 
         & AP@0.1 & AP@0.3 & AP@0.5 & AP@0.0 & AP@0.2 & AP@0.4 & \\ \hline
         Short & .411 & .346 & .235 & .149 & .140 & .051 & .357 \\ \hline
Medium & .675 & .671 & .623& .476 & .361 & .156 & .435 \\ \hline
Long & .781 & .703 & .420 & .704 & .362& .130 & .393 \\ \hline
    \end{tabular}
\end{table}

\section{Speed test}

The inference time per video clip is shown in Table~\ref{tab:inference_time}. The speed test is conducted on one Titan X GPU.  Inference times for all models are under 1 second. Baselines 1 and 2 are faster as the model architecture is simpler. Our model takes roughly twice the time of baseline 3, which is mainly due to the second-stage refinement. 
\begin{table}[h]
  \caption{\label{tab:inference_time} Inference time per 300-frame video clip}
  \begin{tabular}{c|cccc}\hline
    & Base 1 & Base 2 & Base 3 & Ours \\ \hline
    Inference time (ms) & 10.9 & 11.6 & 284.5 & 511.1 \\ \hline
  \end{tabular}
\end{table}

\section{Detection examples\bs{}}
\begin{figure*}[t]
  \caption{\label{fig:examples}Detection examples. Red: ground-truth segment, green: predicted segment. The \kledit{sequences are} downsampled.}
  \begin{minipage}{\textwidth}
    \includegraphics[width=\linewidth]{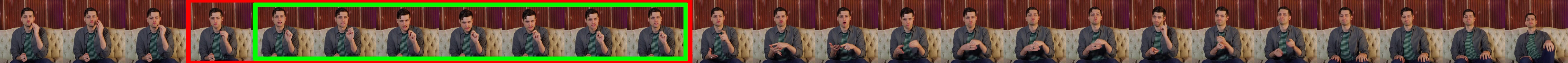}
  \end{minipage}
  \vspace{0.1in}

  \begin{minipage}{\textwidth}
    \includegraphics[width=\linewidth]{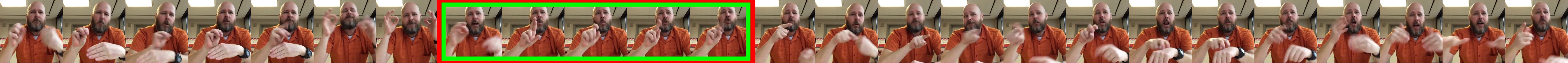}
  \end{minipage}
  \vspace{0.1in}

  \begin{minipage}{\textwidth}
    \includegraphics[width=\linewidth]{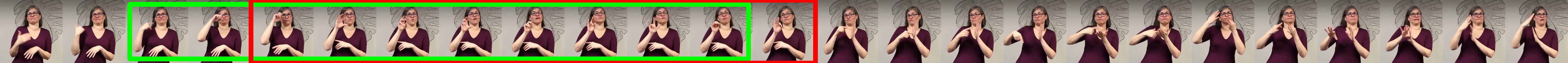}
  \end{minipage}
  \vspace{0.1in}

  \begin{minipage}{\textwidth}
    \includegraphics[width=\linewidth]{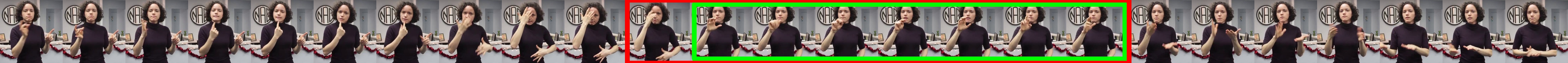}
  \end{minipage}
  \vspace{0.1in}

  \begin{minipage}{\textwidth}
    \includegraphics[width=\linewidth]{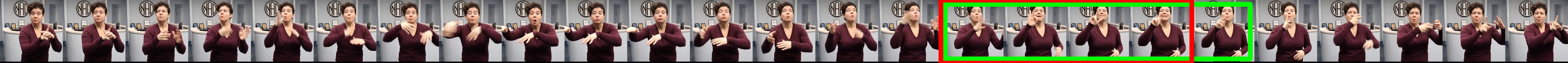}
  \end{minipage}
  \vspace{0.1in}

  \begin{minipage}{\textwidth}
    \includegraphics[width=\linewidth]{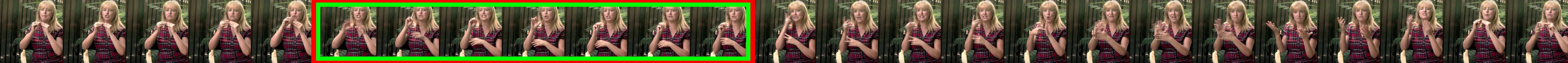}
  \end{minipage}
  \vspace{0.1in}

  \begin{minipage}{\textwidth}
    \includegraphics[width=\linewidth]{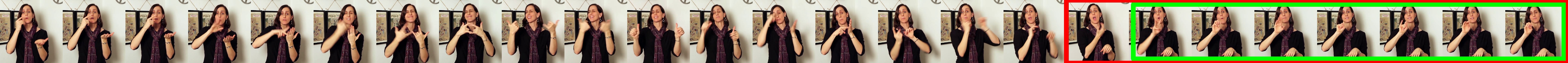}
  \end{minipage}
  \vspace{0.1in}

  \begin{minipage}{\textwidth}
    \includegraphics[width=\linewidth]{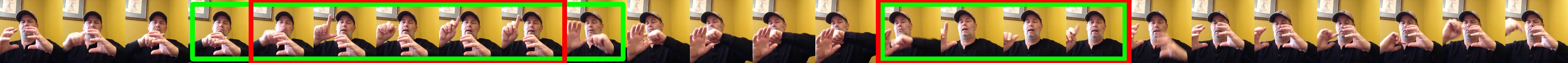}
  \end{minipage}
  \vspace{0.1in}

  \begin{minipage}{\textwidth}
    \includegraphics[width=\linewidth]{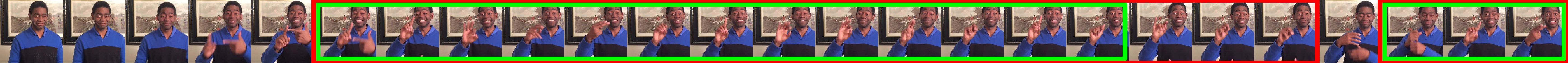}
  \end{minipage}
  \vspace{0.1in}

  \begin{minipage}{\textwidth}
    \includegraphics[width=\linewidth]{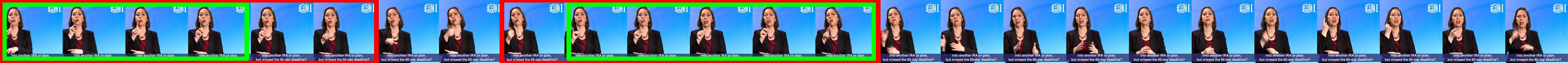}
  \end{minipage}
  \vspace{0.1in}

  \begin{minipage}{\textwidth}
    \includegraphics[width=\linewidth]{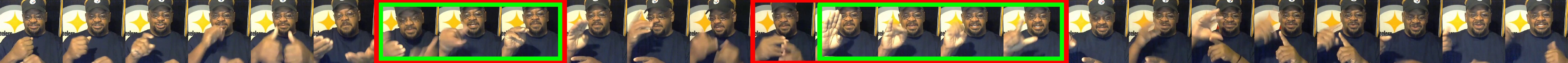}
  \end{minipage}
  \vspace{0.1in}

  \begin{minipage}{\textwidth}
    \includegraphics[width=\linewidth]{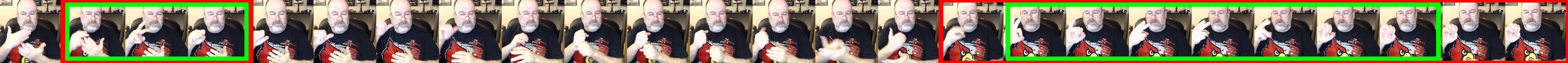}
  \end{minipage}
  \vspace{0.1in}

  \begin{minipage}{\textwidth}
    \includegraphics[width=\linewidth]{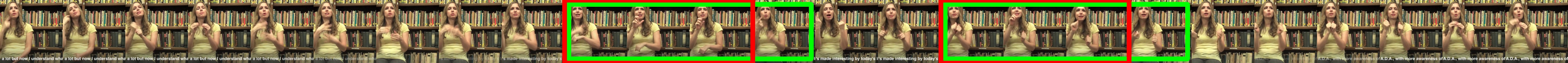}
  \end{minipage}
  \vspace{0.1in}

  \begin{minipage}{\textwidth}
    \includegraphics[width=\linewidth]{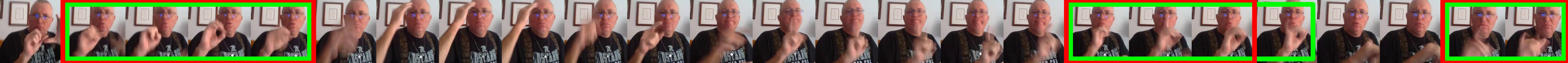}
  \end{minipage}
  \vspace{0.1in}

\end{figure*}

Figure~\ref{fig:examples} shows various detection examples from the ChicagoFSWild dev set. 

\end{document}